\documentclass[conference]{IEEEtran}
\IEEEoverridecommandlockouts
\usepackage{cite}
\usepackage{amsmath,amssymb,amsfonts}
\usepackage{algorithmic}
\usepackage{graphicx}
\usepackage{textcomp}
\usepackage{xcolor}
\usepackage{svg}
\usepackage{subfigure}
\usepackage{url}
\usepackage{paralist}
\usepackage{tikz}
\usepackage{fancyhdr}
\usepackage{afterpage}

\def\BibTeX{{\rm B\kern-.05em{\sc i\kern-.025em b}\kern-.08em
    T\kern-.1667em\lower.7ex\hbox{E}\kern-.125emX}}
\begin{document}

\title{Frontiers of Deep Learning: From Novel Application to Real-World Deployment\\
}

\author{\IEEEauthorblockN{Rui Xie}
\IEEEauthorblockA{
Department of Electrical, Computer, and Systems Engineering\\Rensselaer Polytechnic Institute\\
xier2@rpi.edu}
}

\maketitle
\thispagestyle{fancy}

\begin{abstract}
Deep learning continues to re-shape numerous fields, from natural language processing and imaging to data analytics and recommendation systems. This report studies two research papers that represent recent progress on deep learning from two largely different aspects: The first paper applied the transformer networks, which are typically used in language models, to improve the quality of synthetic aperture radar image by effectively reducing the speckle noise. The second paper presents an in-storage computing design solution to enable cost-efficient and high-performance implementations of deep learning recommendation systems. In addition to summarizing each paper in terms of motivation, key ideas and techniques, and evaluation results, this report also presents thoughts and discussions about possible future research directions. By carrying out in-depth study on these two representative papers and related references, this doctoral candidate has developed better understanding on the far-reaching impact and efficient implementation of deep learning models. 
\end{abstract}

\begin{IEEEkeywords}
Deep learning, Synthetic aperture radar, Transformer, In-storage computing, Architecture
\end{IEEEkeywords}

\section{Introduction}
This report presents our study on two papers that represent important accomplishments in two distinct areas related to deep learning. The first paper~\cite{perera2022transformer} is about applying recent advancement of deep learning models to benefit synthetic aperture radar~(SAR) imaging. Deep learning signals a promising shift in how we perceive, interpret and manipulate visual data. The recent advent of transformer-based architectures brings exciting opportunities to improve the state of the art of imaging technologies. Characterized by their self-attention mechanisms, these architectures have shown intriguing capabilities to discern intricate patterns within data. This has led to novel applications of transformers in various imaging systems, such as remote sensing~\cite{perera2022transformer, du2022transunet++, zhou2022pvt, liu2022high}, medical imaging~\cite{chen2021transunet,cao2022swin,hatamizadeh2022unetr} and image restoration~\cite{liang2021swinir,zamir2022restormer,wang2022uformer}. This well exemplifies the versatility and potential harbored by transformer architectures in addressing diverse challenges inherent to imaging domains. 

The second paper~\cite{sun2022rm} aims at improving deep learning system implementation efficiency via in-storage computing. The proliferation of large-scale deep learning models presents grand challenges to meet the computing and memory demands at affordable implementation cost, which makes wide deployment of deep learning a non-trivial endeavor. This is especially critical for real-time applications where latency is an important factor, or edge devices where computational resources can be severely constrained~\cite{patwardhan2023transformers,bondarenko2021understanding,reidy2023work,lumenEdgeComputing}. As a result, significant research efforts~(e.g., see~\cite{fbTransparentMemory, micronHBM3Memory, shen2023efficient, choi2021efficient, frontiersinTransformComputationally, chen2020deep} have been devoted to developing highly efficient deep learning implementation platforms. Heterogeneous computing with computational task offloading has been considered as a viable pathway to ameliorate the computational burden. Offloading essentially entails the redistribution of computational tasks into purpose-built, low-cost devices, thereby optimizing the utilization of available resources, maintaining low latency, and enabling the deployment of sophisticated deep learning models in a wider array of settings~\cite{infoqMicrosoftReleases, beaumont2020optimal,um2023fastflow,zabihi2023reinforcement,fbTransparentMemory,chen2022general}. 

Through studying these two papers and relevant references, this doctoral candidate has developed a deeper understanding on the potential of applying deep learning to improve imaging applications and applying novel near-data computing to improve deep learning system implementation efficiency. This candidate sincerely appreciate the committee choosing these two important and inspiring papers. The rest of this report is organized as follows: Section~\ref{sec:sar image despeckling} summarizes the study on the first paper~\cite{perera2022transformer} with additional thoughts on the future research directions in the area of deep learning for imaging technologies.  Section~\ref{sec:rm-ssd} summarize the study on the second paper~\cite{sun2022rm} with additional thoughts on the future research directions in the area of efficient deep learning implementation. Further discussions and conclusions are drawn in Section~\ref{sec:discussion and conclusion}.

\begin{figure*}[t]
\centering 
\includegraphics[width=.9\linewidth]{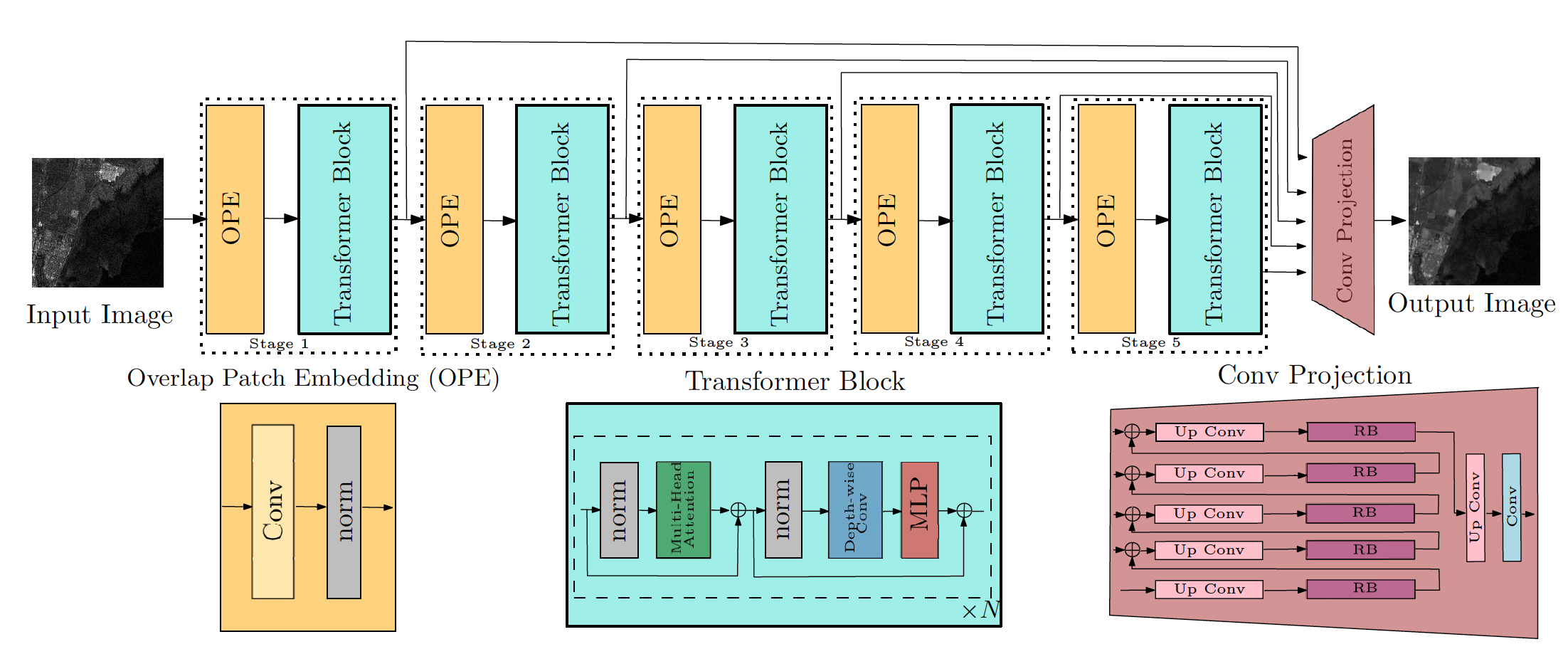}
\caption{Overview of transformer-based despeckling transformer architecture~\cite{perera2022transformer}.} 
\label{fig:network overview} 
\end{figure*}
\section{Transformer-Based Architecture for SAR Image Despeckling} \label{sec:sar image despeckling}

\subsection{Background}
The first paper revealed the potential of applying transformer-based architecture to improve SAR despeckling performance. Prior work~\cite{chierchia2017sar, wang2017sar, zhu2021deep} has shown that convolutional neural networks~(CNNs) could surpass traditional methods on SAR despeckling performance. However, CNN-based methods still demand the use of both noisy and clean (ground truth) images for supervised training. Given the inherent absence of clean SAR references, these methods often utilize synthetic speckle generation or multi-temporal fusion to generate the training sets.

Recent success of transformers, especially in natural language processing, has stimulated its wide applications in computer vision tasks. Vision Transformer~(ViT)~\cite{dosovitskiy2020image} showcased remarkable image classification prowess on ImageNet~\cite{deng2009imagenet} by capitalizing on the ability of transformers to discern long-range dependencies in images. 
The first paper that we studied presents pioneering work that utilizes transformer-based architecture in SAR despeckling.

\subsection{Problem Statement}
SAR imaging naturally has speckle noise, which makes the image look grainy. This graininess comes from the way that radar signals are processed, and can hide important details in the image, making it hard to analyze. This work aims to use transformers to improve the clarity and preserve the details in SAR images.

Speckle noise in SAR images is of a multiplicative nature. For an SAR image with an average number of \(L\) looks, the observed SAR intensity \(y\) is related to the speckle-free SAR intensity \(x\) as follows:
\begin{equation}
y = x \cdot n,
\end{equation}
where \(n\) represents the multiplicative speckle. Assuming fully developed speckle, \(n\) follows a Gamma distribution with unit mean and a variance of \(1/L\). The probability distribution of \(n\) can be expressed as:
\begin{equation}
p(n) = \frac{1}{\Gamma(L)} L^L n^{L-1} e^{-Ln},
\end{equation}
where \(\Gamma(\cdot)\) denotes the Gamma function. Given \(y\), the primary goal here is to estimate \(x\).

\subsection{Proposed Design Solution}
\subsubsection{Network Architecture}
The proposed network architecture, depicted in Fig.~\ref{fig:network overview}. The input image $y$ passes through an encoder with five stages, each containing an overlap patch embedding block followed by a transformer block. This hierarchical setup in the encoder captures both the high-resolution fine features and low-resolution coarse features, which are important for effective SAR despeckling. We further elaborate on the key architectural components as follows.

\paragraph{Overlap Patch Embedding Block}
The objective of this block is to supply the transformer block with the overlapping feature patches. The input first processes through a convolutional layer with the following specifications: kernel size \(k\), embedded dimensions \(e\), stride \(s\), and padding \(p = k/2\). Accordingly, layer normalization is applied to the output for stable and better training performance.

\paragraph{Transformer Block}
SAR despeckling requires capturing both the local and global dependencies within the image. While traditional convolutional networks are proficient in encoding local patterns, they might fall short in handling long-range dependencies. The transformer block in the proposed architecture addresses this limitation with self-attention mechanism, which assesses the importance of different regions in the image, irrespective of their spatial closeness~\cite{guo2022transformer,pongrac2023despeckling}.

The transformer block is formulated as:
\begin{equation}
T(\textbf{I}_{\text{in}}) = \text{MLP}(\text{DWC}(X(\textbf{I}_{\text{in}}))) + X(\textbf{I}_{\text{in}}), 
\end{equation}
where we define $X(\textbf{I}_{\text{in}}) = \text{MHA}(\textbf{I}_{\text{in}}) + \textbf{I}_{\text{in}}$, MHA denotes  the multi-head attention layer, DWC denotes the depth-wise convolution, and MLP denotes the multi-layer perceptron. Layer normalization is performed on \(I_{\text{in}}\) and \(X(I_{\text{in}})\) before being passed to the MHA and DWC, respectively.

The block employs multiple attention heads
to enhance the model's capacity to attend to different features concurrently. To improve computational efficiency, a reduction ratio $R$ is incorporated to lower the spatial dimensions before performing the attention computation~\cite{wang2021pyramid}.

\paragraph{Convolutional Projection Block}
This block is responsible for upsampling the outputs from the transformer blocks to align with the original image dimensions. The upsampling layers enhance the resolution by a factor of two, ensuring the spatial dimensions of the despeckled output match the input. The Residual Block (RB) for a given input \(I_{\text{in}}\) can be expressed as:
\begin{equation}
\text{RB}(\textbf{I}_{\text{in}}) = \text{Conv}_{3 \times 3}(\text{ReLU}(\text{Conv}_{3 \times 3}(\textbf{I}_{\text{in}}))) + \textbf{I}_{\text{in}},
\end{equation}
where \(\text{Conv}_{3 \times 3}\) represents a \(3 \times 3\) convolution layer and ReLU denotes the rectified linear unit activation function.

\subsubsection{Loss Function}
In order to strike an appropriate balance between reducing the speckle noise and preserving the essential details of the image, the proposed method combines two loss functions: 
The primary loss employed is the \(l_2\) loss, defined as:
\begin{equation}
L_{l_2} = ||x - \hat{x}||_2^2,
\end{equation}
which focuses on minimizing the error between the ground truth \(x\) and the predicted output \(\hat{x}\), driving the network to produce despeckled images. However, merely minimizing the \(l_2\) loss can lead to over-smoothed predictions. To counteract this and ensure the preservation of image details, especially edges, a total variation loss is introduced:
\begin{equation}
L_{tv} = \sum_{i, j} | \hat{x}_{i+1, j} - \hat{x}_{i, j} | + | \hat{x}_{i, j+1} - \hat{x}_{i, j} |.
\end{equation}
The combined objective function is a weighted sum of these two losses:
\begin{equation}
L = \lambda_1 L_{l_2} + \lambda_2 L_{tv},
\end{equation}
where the weights \(\lambda_1\) and \(\lambda_2\) are chosen to balance the contributions of speckle reduction and detail preservation.

\subsection{Evaluation}
The paper carried out extensive experiments to evaluate the proposed design solution using both synthetic images of Set12 dataset~\cite{zhang2017beyond} and real SAR images. On synthetic images, the proposed method achieved a PSNR~(peak signal-to-noise ratio) of 24.56, which is significantly better than conventional practice (e.g., PPB~\cite{deledalle2009iterative} with PSNR of 21.90) and slightly better than SAR-CNN~\cite{chierchia2017sar}'s 24.51. In terms of SSIM~(structured similarity index), the proposed design solution scores 0.718, which is noticeably better than SAR-CNN's 0.651. When tested on real SAR images, the proposed design solution shows similar results, outperforming all other competing methods.

\subsection{Further Thoughts}
The paper well demonstrated the effectiveness of combining the new  transformer architecture and CNN to improve SAR image despeckling. We expect that the research community will continue to push the SAR image despeckling performance envelope by applying more advanced and larger-scale neural network models. Nevertheless, we believe that the research community should also pay more attention to the practical  deployment of the solutions, especially in real-time scenarios~(e.g., disaster response, real-time monitoring, navigation), in the presence of the computationally intensive nature of transformers~\cite{li2023transforming,ansar2023novel,wu2022towards,wang2022matchformer}. Accordingly, we would like to list some thoughts on potential future research directions:
\begin{compactenum}
    \item {\it Hybrid Architectures:}~A potential avenue for improvement could be exploring hybrid architectures that meld transformer blocks for global feature extraction with convolutional blocks for local (spatial) feature extraction. This could potentially offer a balanced approach to effectively extract and enhance useful features from SAR images while improving the implementation efficiency.
    \item {\it Hardware Acceleration:} Recent research~\cite{zhou2022transpim,li2017drisa, gao2017tetris} has demonstrated the power of software-hardware co-design to significantly improve the deep learning system implementation efficiency. This could be future customized towards the remote sensing area. In particular, we expect that sparse transformer models~\cite{fang2022efficient} could be attractive in this context.
    \item {\it Automated Hyperparameter Tuning and Model Optimization:} Optimizing the hyperparameters can greatly influence deep learning system performance and computational efficiency. Automated tuning frameworks such as OptFormer~\cite{chen2022towards} could be employed to systematically search for optimal hyperparameters, which can enhance the performance and efficiency in processing SAR images while reducing the manual effort required in the tuning process~\cite{shawki2021automating}.
\end{compactenum}


\section{Efficient Implementation via In-Storage Computing} \label{sec:rm-ssd}

\subsection{Background}
The second paper presents a solution that can improve recommendation system implementation efficiency via in-storage computing. Recommendation systems have become an important part of various online services. Their primary purpose is to offer personalized recommendations based on individual user preferences and interests. Recent progress of deep neural network~(DNN) has led to significant interest on DNN-based recommendation models. Apart from conventional multi-layer perceptron~(MLP) layer, such models often incorporate a data-intensive embedding layer to manage the sparse categorical input features, which can span from tens of millions to even tens of billions of possible categories. In the embedding layer, these sparse features are mapped to embedding vectors~(EVs) and stored across different embedding tables. During recommendation inference, the relevant EVs are fetched based on lookup indices, a process termed as embedding lookup. Subsequently, vectors from each embedding table are aggregated and then passed to the MLP layers for further processing.

\subsection{Problem Statement}
In conventional implementation, the entire set of embeddings is stored in the DRAM to minimize the service latency. As the embedding table size continues to increase~(e.g., from tens of GBs to hundreds of GB or even TBs), it has become more and more difficult to keep all the embedding tables entirely inside DRAM. This results in a significant bottleneck for deploying advanced recommendation systems. To address these challenges, innovative solutions are needed that can leverage low-cost memory technologies~(e.g., SSD) and computational strategies while ensuring that recommendation systems remain efficient, fast, and accurate.

\begin{figure}[t]
\centering 
\includegraphics[width=\linewidth]{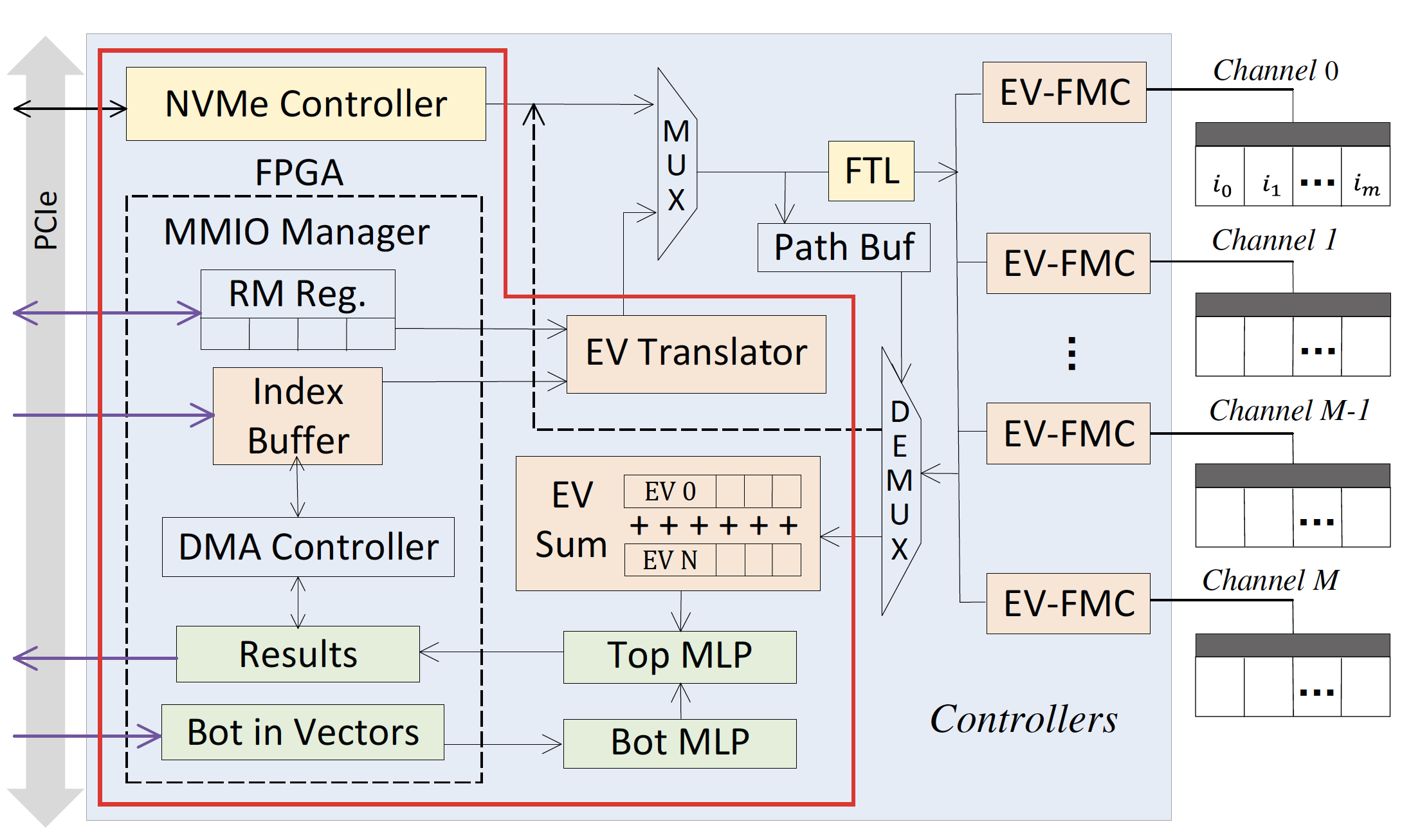}
\caption{Illustration of the proposed RM-SSD~\cite{sun2022rm} that contains: (1){\it Conventional components} (yellow blocks) for conventional SSD control; (2) {\it Embedding lookup engine} (orange blocks) for fast retrieval of embeddings; (3) {\it MLP acceleration engine} (green blocks) for handling the MLP part of the recommendation model.} 
\label{fig:rm-ssd overview} 
\end{figure}

\subsection{Proposed Design Solution}
RM-SSD~\cite{sun2022rm} is an in-storage computing system that realizes SSD-based recommendation inference. The design is motivated by the need to improve efficiency and performance in recommendation systems, particularly when handling large-scale datasets. As shown in Fig.~\ref{fig:rm-ssd overview}, RM-SSD contains three major parts, including {\it conventional components}, {\it embedding lookup engine}, and {\it MLP acceleration engine}.

\subsubsection{Conventional Components} They include NVMe controller for host-SSD communication, flash translation layer~(FTL) for address translation, and flash memory controllers~(FMCs) for managing/accessing all the flash memory chips. The {\it MMIO manager} handles the I/O stack and supports direct access for quick parameter exchanges and DMA~(direct memory access) mode for bulk data transfers.



\subsubsection{Embedding Lookup Engine} It minimizes the latency of accessing embedding vectors~(EVs) by exploiting the intra-SSD data access parallelism. It consists of an {\it EV translator}, an {\it EV-FMC}~(EV flash memory controller), and an {\it EV sum} for EV aggregations. The read process consists of (a) device channel stage during which the {\it EV translator} interprets embedding lookup indices and the {\it EV sum} aggregates the EVs fetched by the EV-FMC; (b) flash channel stage during which EV-FMC retrieves the necessary EVs from flash channels.

\paragraph{EV Translator}
It maps the EV index to logical block addresses~(LBAs), using system calls to retrieve file LBA details for each embedding table. This information, combined with the fixed dimension of each EV, facilitates the computation of index ranges for each extent. These extents and table IDs are stored in DRAM on the FPGA~(field programmablle gate array) inside SSD for fast access.

\paragraph{EV-FMC} 
It aims for efficient retrieval of EVs by dispersing read requests uniformly across all the flash memory channels and dies. After FTL translation, these requests are directed to the corresponding EV-FMC. A multiplexer (MUX) is integrated to prioritize data requests, and paths of read requests are logged in the path buffer for data coalescence. This setup minimizes the data transfer time and boosts the throughput.

\paragraph{EV Sum} 
This component manages embedding vector reads by using a demultiplexer (DEMUX) to differentiate standard block I/O from embedding vector reads. It employs FPGA-based floating-point adders to aggregate embedding vectors efficiently. Its results are sent to the {\it MLP acceleration engine}, and each table yields a vector, with the combined size being \(EV_{dim} \cdot M\).

\subsubsection{MLP Acceleration Engine} This engine is fine-tuned to most effectively implement MLP computation by exploiting the recommendation model's characteristics to ensure the hardware resource utilization efficiency. 

\paragraph{FC Layer} An adder tree approach is employed for the kernel block sum operation of matrix multiplication to curtail the time cost. The design retains MLP weights and embedding vectors in FP32 precision to ensure data integrity.

\paragraph{Intra-Layer Decomposition} Recommendation models inherently involve a feature interaction operation, notably the concatenation, which occurs post the bottom MLP and embedding layer. FPGA has abundant hardware resources that enable highly parallel operations, which can be leveraged to improve MLP processing. By determining the weights of the concatenation that map to the first FC layer $L_0$ of the top MLP, the representation \( RC \) is decomposed into \( R_bC + R_eC \), given an FC layer with $R$ inputs and $C$ outputs. This decomposition allows both the bottom MLP and embedding layer to process simultaneously. Such a concurrent approach ensures that neither the embedding layer nor the bottom MLP becomes a performance bottleneck for $L_0$, allowing the system to achieve a higher throughput and lower latency.

\begin{figure}[htbp]
\centering 
\includegraphics[width=\linewidth]{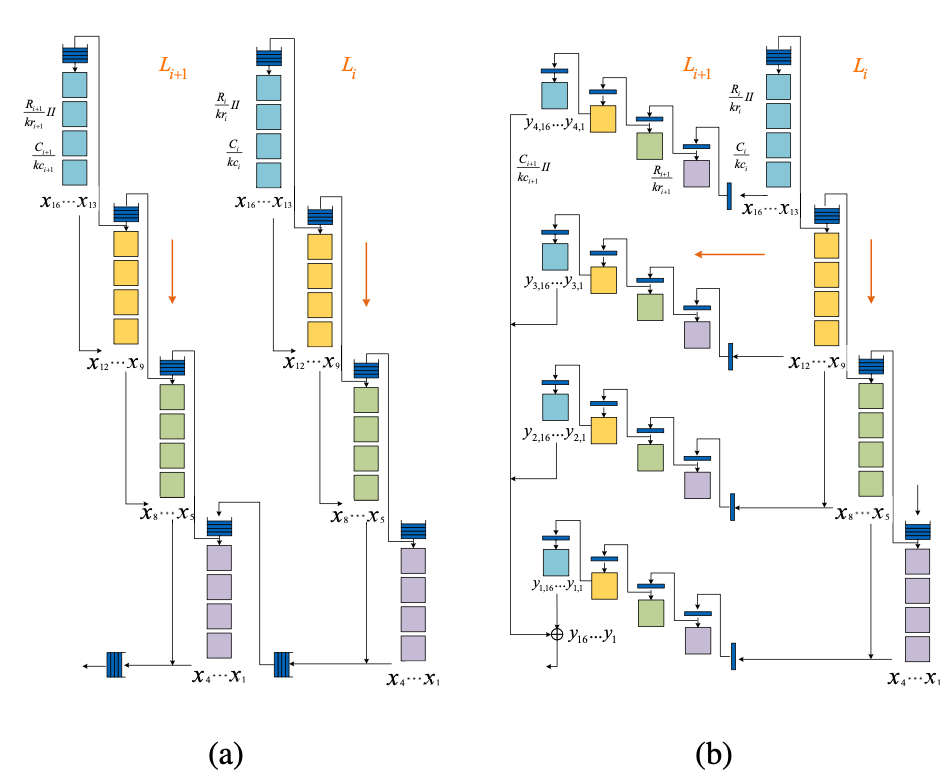}
\caption{Illustration of (a) conventional design practice where a layer $L_{i+1}$ can only receive the first input vector after all columns of $L_i$ have completed accumulation, and (b) proposed inter-layer composition that  implements row scanning in $L_{i+1}$ while maintaining column scanning in $L_i$.} 
\label{fig:alternative scan} 
\end{figure}

\paragraph{Inter-Layer Composition} 
To further improve the processing performance and exploit the FPGA's parallel computational resources, this work introduces an alternating scanning direction between adjacent layers, as illustrated in Fig.~\ref{fig:alternative scan}. While the layer \( L_{i} \) processes the column scanning, the layer \( L_{i+1} \) can concurrently utilizes the row scanning. This strategic processing alteration effectively halves the MLP's processing time, leading to a higher performance.

\paragraph{Kernel Search Algorithm} 
The matrix multiplication kernel function largely dominates the hardware resource utilization. This algorithm finds a balance between embedding-centric and MLP-centric models, which determines the optimal kernel size for each FC layer. This can reduce the hardware resource usage while ensuring a high  throughput. Considering the time expens associated with the embedding layer, bottom MLP, and top MLP, the optimization task can be formulated as follows: search for the most suitable kernel pairs $K_{\text{bot'}} = \{(kr_0, kc_0), (kr_1, kc_1), \ldots, (kr_i, kc_i)\}$ and $K_{\text{top'}} = \{(kr_1, kc_1), \ldots, (kr_j, kc_j)\}$
subject to the following constraints:
\begin{equation} \label{equ:kernel search}
\left\{
\begin{aligned}
& T_{\text{bot'}} \leq T_{\text{emb'}} \\
& T_{\text{top'}} \leq T_{\text{emb'}} \\
& \mathop\text{argmin}\limits_{K_{\text{bot'}}, K_{\text{top'}}} \sum_i kr_{i}kc_{i} + \sum_j kr_j kc_j + kr_e kc_e
\end{aligned}
\right.
\end{equation}
Where $T_{\text{bot'}}, T_{\text{top'}}, T_{\text{emb'}}$ are time cost for bottom MLP, top MLP and embedding layer, $kr, kc$ are kernel block sizes along row and column dimension.
Several guiding rules can be used to facilitate the optimization: (i)~{\textit{BRAM resource assessment:} Utilize FPGA block RAM~(BRAM) for weight storage primarily. If combined weights exceed the BRAM capacity, resort to off-chip DRAM. (ii)~\textit{Kernel size for DRAM employed layers:} When DRAM serves an FC layer, FPGA retains the weights fetched from DRAM until consumed to better utilize the DRAM bandwidth. (iii)~{\textit{Batch size decision:} To improve the implementation efficiency, we should adjust batch size dynamically if certain conditions related to kernel sizes and time costs are not met. (iv)~{\textit{Kernel pair selection:} We should start with the larger kernel pairs and fine-tune to minimize pipeline interruptions, ensuring balanced kernel sizes to prevent any single layer from becoming a bottleneck.

\subsection{Evaluation}
RM-SSD was tested on AWS EC2 F1 Instances~\cite{amazonAmazonInstances}, showcasing remarkable results: When being benchmarked against Facebook's DLRM models~\cite{naumov2019deep}, the embedding lookup engine only model EMB-VectorSum
significantly outperformed the baseline SSD-S~(running with DRAM size equals to 1/4 of embedding tables size)
by 16×. Overall, RM-SSD achieved up to a 36× throughput improvement and a 97\% reduction in latency compared to baseline SSD-S. Notably, when running models such as Neural Collaborative Filtering~(NCF)~\cite{he2017neural} and Wide \& Deep~(WnD)~\cite{cheng2016wide}, RM-SSD can outperform the baseline SSD-S by around 100×. Furthermore, using an optimized kernel search algorithm, RM-SSD maintained performance with significantly reduced FPGA resource consumption. Specifically, when running the meta recommendation model RMC3~\cite{naumov2019deep}, the proposed design solution can reduce the usage of LUTs, FFs, BRAM, and DSPs by 40\%, 40\%, 10\%, and 40\%, respectively.
The evaluation results well demonstrated that RM-SSD is a highly resource-efficient solution for recommendation systems.

\subsection{Further Thoughts}
In-storage computing, facilitated by SSD's capacity and bandwidth, arises as a promising solution to better serve growing data-intensive applications such as AI and data analytics. This has been well demonstrated by RM-SSD in the context of recommendation systems. Most prior work on in-storage computing (including RM-SSD) assumed that intra-SSD flash memory access bandwidth is (much) higher than the SSD I/O interface bandwidth, which however may become invalid as the PCIe bandwidth steadily increases. This could warrant re-evaluation of the effectiveness of many proposed design solutions. Moreover, to further improve the effectiveness, future research may be directed to the following aspects:
\begin{compactenum}
    \item{\textit{Scalability and programmability}}: RM-SSD specializes at recommendation systems and demands non-trivial system integration efforts. To ensure wide-range deployment of in-storage computing in the real world, highly scalable and programmable frameworks are indispensable. This demands significant amount of future research, in spite of some recent studies~\cite{mailthody2019deepstore, gu2016biscuit}. Such frameworks should seamlessly support high-level programming interface and support a wide variety of AI models. 
    \item{\textit{Resource-efficient algorithms}}: Given limited computing resources inside storage devices, it is highly desirable to develop software-hardware optimization techniques (e.g., see~\cite{shi2023efficient,colbert2021competitive,liu2022hybridc}) that can better utilize hardware resources for various AI systems.
    \item{\textit{Interference with flash memory management}}: For storage devices with in-storage computing capability, in addition to serve the computing tasks, they must also serve normal I/O requests. These two duties can compete for the intra-SSD DRAM and flash memory bandwidth, leading to interference to each other. Future research should be conducted to develop techniques that can most appropriately schedule these two duties to minimize the interference and hence maximize the overall system performance. 
\end{compactenum}


\section{Conclusion}\label{sec:discussion and conclusion}
This report presents our studies of the two papers in the areas of transformer-assisted SAR despeckling and recommendation system acceleration. For each paper, we summarize its motivation, present its key idea and design techniques, and highlight the evaluation results. Moreover, by reading additional references to better understand the state of the art in each area, we list some thoughts about possible future research directions. By carrying out this study and writing this report, this doctoral candidate has gained valuable further understanding of deep learning and its broader impact. In conclusion, deep learning has made extraordinary progress over the past decade, and future research will undoubtedly continue to improve deep learning from new algorithms and applications to efficient hardware/software implementations. 

\bibliographystyle{IEEEtran}
\bibliography{ref}

\begin{thebibliography}{10}
\providecommand{\url}[1]{#1}
\csname url@samestyle\endcsname
\providecommand{\newblock}{\relax}
\providecommand{\bibinfo}[2]{#2}
\providecommand{\BIBentrySTDinterwordspacing}{\spaceskip=0pt\relax}
\providecommand{\BIBentryALTinterwordstretchfactor}{4}
\providecommand{\BIBentryALTinterwordspacing}{\spaceskip=\fontdimen2\font plus
\BIBentryALTinterwordstretchfactor\fontdimen3\font minus \fontdimen4\font\relax}
\providecommand{\BIBforeignlanguage}[2]{{%
\expandafter\ifx\csname l@#1\endcsname\relax
\typeout{** WARNING: IEEEtran.bst: No hyphenation pattern has been}%
\typeout{** loaded for the language `#1'. Using the pattern for}%
\typeout{** the default language instead.}%
\else
\language=\csname l@#1\endcsname
\fi
#2}}
\providecommand{\BIBdecl}{\relax}
\BIBdecl

\bibitem{perera2022transformer}
M.~V. Perera, W.~G.~C. Bandara, J.~M.~J. Valanarasu, and V.~M. Patel, ``Transformer-based sar image despeckling,'' in \emph{IGARSS IEEE International Geoscience and Remote Sensing Symposium}.\hskip 1em plus 0.5em minus 0.4em\relax IEEE, 2022, pp. 751--754.

\bibitem{du2022transunet++}
Y.~Du, R.~Zhong, Q.~Li, and F.~Zhang, ``Transunet++ sar: Change detection with deep learning about architectural ensemble in sar images,'' \emph{Remote Sensing}, vol.~15, no.~1, p.~6, 2022.

\bibitem{zhou2022pvt}
Y.~Zhou, X.~Jiang, G.~Xu, X.~Yang, X.~Liu, and Z.~Li, ``Pvt-sar: An arbitrarily oriented sar ship detector with pyramid vision transformer,'' \emph{IEEE Journal of Selected Topics in Applied Earth Observations and Remote Sensing}, vol.~16, pp. 291--305, 2022.

\bibitem{liu2022high}
X.~Liu, Y.~Wu, W.~Liang, Y.~Cao, and M.~Li, ``High resolution sar image classification using global-local network structure based on vision transformer and cnn,'' \emph{IEEE Geoscience and Remote Sensing Letters}, vol.~19, pp. 1--5, 2022.

\bibitem{chen2021transunet}
J.~Chen, Y.~Lu, Q.~Yu, X.~Luo, E.~Adeli, Y.~Wang, L.~Lu, A.~L. Yuille, and Y.~Zhou, ``Transunet: Transformers make strong encoders for medical image segmentation,'' \emph{arXiv preprint arXiv:2102.04306}, 2021.

\bibitem{cao2022swin}
H.~Cao, Y.~Wang, J.~Chen, D.~Jiang, X.~Zhang, Q.~Tian, and M.~Wang, ``Swin-unet: Unet-like pure transformer for medical image segmentation,'' in \emph{European conference on computer vision}.\hskip 1em plus 0.5em minus 0.4em\relax Springer, 2022, pp. 205--218.

\bibitem{hatamizadeh2022unetr}
A.~Hatamizadeh, Y.~Tang, V.~Nath, D.~Yang, A.~Myronenko, B.~Landman, H.~R. Roth, and D.~Xu, ``Unetr: Transformers for 3d medical image segmentation,'' in \emph{Proceedings of the IEEE/CVF winter conference on applications of computer vision}, 2022, pp. 574--584.

\bibitem{liang2021swinir}
J.~Liang, J.~Cao, G.~Sun, K.~Zhang, L.~Van~Gool, and R.~Timofte, ``Swinir: Image restoration using swin transformer,'' in \emph{Proceedings of the IEEE/CVF international conference on computer vision}, 2021, pp. 1833--1844.

\bibitem{zamir2022restormer}
S.~W. Zamir, A.~Arora, S.~Khan, M.~Hayat, F.~S. Khan, and M.-H. Yang, ``Restormer: Efficient transformer for high-resolution image restoration,'' in \emph{Proceedings of the IEEE/CVF conference on computer vision and pattern recognition}, 2022, pp. 5728--5739.

\bibitem{wang2022uformer}
Z.~Wang, X.~Cun, J.~Bao, W.~Zhou, J.~Liu, and H.~Li, ``Uformer: A general u-shaped transformer for image restoration,'' in \emph{Proceedings of the IEEE/CVF conference on computer vision and pattern recognition}, 2022, pp. 17\,683--17\,693.

\bibitem{sun2022rm}
X.~Sun, H.~Wan, Q.~Li, C.-L. Yang, T.-W. Kuo, and C.~J. Xue, ``Rm-ssd: In-storage computing for large-scale recommendation inference,'' in \emph{IEEE International Symposium on High-Performance Computer Architecture (HPCA)}.\hskip 1em plus 0.5em minus 0.4em\relax IEEE, 2022, pp. 1056--1070.

\bibitem{patwardhan2023transformers}
N.~Patwardhan, S.~Marrone, and C.~Sansone, ``Transformers in the real world: A survey on nlp applications,'' \emph{Information}, vol.~14, no.~4, p. 242, 2023.

\bibitem{bondarenko2021understanding}
Y.~Bondarenko, M.~Nagel, and T.~Blankevoort, ``Understanding and overcoming the challenges of efficient transformer quantization,'' \emph{arXiv preprint arXiv:2109.12948}, 2021.

\bibitem{reidy2023work}
B.~Reidy, M.~Mohammadi, M.~Elbtity, H.~Smith, and Z.~Ramtin, ``Work in progress: Real-time transformer inference on edge ai accelerators,'' in \emph{IEEE 29th Real-Time and Embedded Technology and Applications Symposium (RTAS)}.\hskip 1em plus 0.5em minus 0.4em\relax IEEE, 2023, pp. 341--344.

\bibitem{lumenEdgeComputing}
R.~Tucker, ``{E}dge {C}omputing {T}ransforms {B}usinesses with {L}ow {L}atency, {R}eal-{T}ime {D}ata and {C}ompute {P}rovisioning,'' \url{https://blog.lumen.com/}, [Accessed 18-10-2023].

\bibitem{fbTransparentMemory}
``{T}ransparent memory offloading: more memory at a fraction of the cost and power --- engineering.fb.com,'' \url{https://engineering.fb.com/2022/06/20/data-infrastructure/}, [Accessed 18-10-2023].

\bibitem{micronHBM3Memory}
``{H}{B}{M}3 {M}emory | {H}{B}{M}3 {G}en2 | {M}icron {T}echnology,'' \url{https://www.micron.com/products/ultra-bandwidth-solutions/hbm3}, [Accessed 18-10-2023].

\bibitem{shen2023efficient}
L.~Shen, Y.~Sun, Z.~Yu, L.~Ding, X.~Tian, and D.~Tao, ``On efficient training of large-scale deep learning models: A literature review,'' \emph{arXiv preprint arXiv:2304.03589}, 2023.

\bibitem{choi2021efficient}
H.~Choi and J.~Lee, ``Efficient use of gpu memory for large-scale deep learning model training,'' \emph{Applied Sciences}, vol.~11, no.~21, p. 10377, 2021.

\bibitem{frontiersinTransformComputationally}
``{T}ransform {C}omputationally {I}ntensive {D}eep {L}earning {M}odels to {L}ow-{C}ost {M}odels,'' \url{https://www.frontiersin.org/research-topics/42574/}, [Accessed 18-10-2023].

\bibitem{chen2020deep}
C.~Chen, P.~Zhang, H.~Zhang, J.~Dai, Y.~Yi, H.~Zhang, and Y.~Zhang, ``Deep learning on computational-resource-limited platforms: a survey,'' \emph{Mobile Information Systems}, pp. 1--19, 2020.

\bibitem{infoqMicrosoftReleases}
``{M}icrosoft {R}eleases {A}{I} {T}raining {L}ibrary {Z}e{R}{O}-3 {O}ffload --- infoq.com,'' \url{https://www.infoq.com/news/2021/04/microsoft-zero3-offload/}, [Accessed 18-10-2023].

\bibitem{beaumont2020optimal}
O.~Beaumont, L.~Eyraud-Dubois, and A.~Shilova, ``Optimal gpu-cpu offloading strategies for deep neural network training,'' in \emph{European Conference on Parallel Processing}.\hskip 1em plus 0.5em minus 0.4em\relax Springer, 2020, pp. 151--166.

\bibitem{um2023fastflow}
T.~Um, B.~Oh, B.~Seo, M.~Kweun, G.~Kim, and W.-Y. Lee, ``Fastflow: Accelerating deep learning model training with smart offloading of input data pipeline,'' \emph{Proceedings of the VLDB Endowment}, vol.~16, no.~5, pp. 1086--1099, 2023.

\bibitem{zabihi2023reinforcement}
Z.~Zabihi, A.~M. Eftekhari~Moghadam, and M.~H. Rezvani, ``Reinforcement learning methods for computation offloading: A systematic review,'' \emph{ACM Computing Surveys}, vol.~56, no.~1, pp. 1--41, 2023.

\bibitem{chen2022general}
D.~Chen, H.~Jin, L.~Zheng, Y.~Huang, P.~Yao, C.~Gui, Q.~Wang, H.~Liu, H.~He, X.~Liao \emph{et~al.}, ``A general offloading approach for near-dram processing-in-memory architectures,'' in \emph{IEEE International Parallel and Distributed Processing Symposium (IPDPS)}.\hskip 1em plus 0.5em minus 0.4em\relax IEEE, 2022, pp. 246--257.

\bibitem{chierchia2017sar}
G.~Chierchia, D.~Cozzolino, G.~Poggi, and L.~Verdoliva, ``Sar image despeckling through convolutional neural networks,'' in \emph{IEEE International Geoscience and Remote Sensing Symposium (IGARSS)}.\hskip 1em plus 0.5em minus 0.4em\relax IEEE, 2017, pp. 5438--5441.

\bibitem{wang2017sar}
P.~Wang, H.~Zhang, and V.~M. Patel, ``Sar image despeckling using a convolutional neural network,'' \emph{IEEE Signal Processing Letters}, vol.~24, no.~12, pp. 1763--1767, 2017.

\bibitem{zhu2021deep}
X.~X. Zhu, S.~Montazeri, M.~Ali, Y.~Hua, Y.~Wang, L.~Mou, Y.~Shi, F.~Xu, and R.~Bamler, ``Deep learning meets sar: Concepts, models, pitfalls, and perspectives,'' \emph{IEEE Geoscience and Remote Sensing Magazine}, vol.~9, no.~4, pp. 143--172, 2021.

\bibitem{dosovitskiy2020image}
A.~Dosovitskiy, L.~Beyer, A.~Kolesnikov, D.~Weissenborn, X.~Zhai, T.~Unterthiner, M.~Dehghani, M.~Minderer, G.~Heigold, S.~Gelly \emph{et~al.}, ``An image is worth 16x16 words: Transformers for image recognition at scale,'' \emph{arXiv preprint arXiv:2010.11929}, 2020.

\bibitem{deng2009imagenet}
J.~Deng, W.~Dong, R.~Socher, L.-J. Li, K.~Li, and L.~Fei-Fei, ``Imagenet: A large-scale hierarchical image database,'' in \emph{IEEE conference on computer vision and pattern recognition}.\hskip 1em plus 0.5em minus 0.4em\relax Ieee, 2009, pp. 248--255.

\bibitem{guo2022transformer}
J.~Guo, N.~Jia, and J.~Bai, ``Transformer based on channel-spatial attention for accurate classification of scenes in remote sensing image,'' \emph{Scientific Reports}, vol.~12, no.~1, p. 15473, 2022.

\bibitem{pongrac2023despeckling}
B.~Pongrac and D.~Gleich, ``Despeckling of sar images using residual twin cnn and multi-resolution attention mechanism,'' \emph{Remote Sensing}, vol.~15, no.~14, p. 3698, 2023.

\bibitem{wang2021pyramid}
W.~Wang, E.~Xie, X.~Li, D.-P. Fan, K.~Song, D.~Liang, T.~Lu, P.~Luo, and L.~Shao, ``Pyramid vision transformer: A versatile backbone for dense prediction without convolutions,'' in \emph{Proceedings of the IEEE/CVF international conference on computer vision}, 2021, pp. 568--578.

\bibitem{zhang2017beyond}
K.~Zhang, W.~Zuo, Y.~Chen, D.~Meng, and L.~Zhang, ``Beyond a gaussian denoiser: Residual learning of deep cnn for image denoising,'' \emph{IEEE transactions on image processing}, vol.~26, no.~7, pp. 3142--3155, 2017.

\bibitem{deledalle2009iterative}
C.-A. Deledalle, L.~Denis, and F.~Tupin, ``Iterative weighted maximum likelihood denoising with probabilistic patch-based weights,'' \emph{IEEE transactions on image processing}, vol.~18, no.~12, pp. 2661--2672, 2009.

\bibitem{li2023transforming}
J.~Li, J.~Chen, Y.~Tang, C.~Wang, B.~A. Landman, and S.~K. Zhou, ``Transforming medical imaging with transformers? a comparative review of key properties, current progresses, and future perspectives,'' \emph{Medical image analysis}, p. 102762, 2023.

\bibitem{ansar2023novel}
W.~Ansar, S.~Goswami, A.~Chakrabarti, and B.~Chakraborty, ``A novel selective learning based transformer encoder architecture with enhanced word representation,'' \emph{Applied Intelligence}, vol.~53, no.~8, pp. 9424--9443, 2023.

\bibitem{wu2022towards}
B.~Wu, J.~Gu, Z.~Li, D.~Cai, X.~He, and W.~Liu, ``Towards efficient adversarial training on vision transformers,'' in \emph{European Conference on Computer Vision}.\hskip 1em plus 0.5em minus 0.4em\relax Springer, 2022, pp. 307--325.

\bibitem{wang2022matchformer}
Q.~Wang, J.~Zhang, K.~Yang, K.~Peng, and R.~Stiefelhagen, ``Matchformer: Interleaving attention in transformers for feature matching,'' in \emph{Proceedings of the Asian Conference on Computer Vision}, 2022, pp. 2746--2762.

\bibitem{zhou2022transpim}
M.~Zhou, W.~Xu, J.~Kang, and T.~Rosing, ``Transpim: A memory-based acceleration via software-hardware co-design for transformer,'' in \emph{IEEE International Symposium on High-Performance Computer Architecture (HPCA)}.\hskip 1em plus 0.5em minus 0.4em\relax IEEE, 2022, pp. 1071--1085.

\bibitem{li2017drisa}
S.~Li, D.~Niu, K.~T. Malladi, H.~Zheng, B.~Brennan, and Y.~Xie, ``Drisa: A dram-based reconfigurable in-situ accelerator,'' in \emph{Proceedings of the 50th Annual IEEE/ACM International Symposium on Microarchitecture}, 2017, pp. 288--301.

\bibitem{gao2017tetris}
M.~Gao, J.~Pu, X.~Yang, M.~Horowitz, and C.~Kozyrakis, ``Tetris: Scalable and efficient neural network acceleration with 3d memory,'' in \emph{Proceedings of the Twenty-Second International Conference on Architectural Support for Programming Languages and Operating Systems}, 2017, pp. 751--764.

\bibitem{fang2022efficient}
C.~Fang, S.~Guo, W.~Wu, J.~Lin, Z.~Wang, M.~K. Hsu, and L.~Liu, ``An efficient hardware accelerator for sparse transformer neural networks,'' in \emph{IEEE International Symposium on Circuits and Systems (ISCAS)}.\hskip 1em plus 0.5em minus 0.4em\relax IEEE, 2022, pp. 2670--2674.

\bibitem{chen2022towards}
Y.~Chen, X.~Song, C.~Lee, Z.~Wang, R.~Zhang, D.~Dohan, K.~Kawakami, G.~Kochanski, A.~Doucet, M.~Ranzato \emph{et~al.}, ``Towards learning universal hyperparameter optimizers with transformers,'' \emph{Advances in Neural Information Processing Systems}, vol.~35, pp. 32\,053--32\,068, 2022.

\bibitem{shawki2021automating}
N.~Shawki, R.~R. Nunez, I.~Obeid, and J.~Picone, ``On automating hyperparameter optimization for deep learning applications,'' in \emph{IEEE Signal Processing in Medicine and Biology Symposium (SPMB)}.\hskip 1em plus 0.5em minus 0.4em\relax IEEE, 2021, pp. 1--7.

\bibitem{amazonAmazonInstances}
``{A}mazon {E}{C}2 {F}1 {I}nstances --- aws.amazon.com,'' \url{https://aws.amazon.com/ec2/instance-types/f1/}, [Accessed 18-10-2023].

\bibitem{naumov2019deep}
M.~Naumov, D.~Mudigere, H.-J.~M. Shi, J.~Huang, N.~Sundaraman, J.~Park, X.~Wang, U.~Gupta, C.-J. Wu, A.~G. Azzolini \emph{et~al.}, ``Deep learning recommendation model for personalization and recommendation systems,'' \emph{arXiv preprint arXiv:1906.00091}, 2019.

\bibitem{he2017neural}
X.~He, L.~Liao, H.~Zhang, L.~Nie, X.~Hu, and T.-S. Chua, ``Neural collaborative filtering,'' in \emph{Proceedings of the 26th international conference on world wide web}, 2017, pp. 173--182.

\bibitem{cheng2016wide}
H.-T. Cheng, L.~Koc, J.~Harmsen, T.~Shaked, T.~Chandra, H.~Aradhye, G.~Anderson, G.~Corrado, W.~Chai, M.~Ispir \emph{et~al.}, ``Wide \& deep learning for recommender systems,'' in \emph{Proceedings of the 1st workshop on deep learning for recommender systems}, 2016, pp. 7--10.

\bibitem{mailthody2019deepstore}
V.~S. Mailthody, Z.~Qureshi, W.~Liang, Z.~Feng, S.~G. De~Gonzalo, Y.~Li, H.~Franke, J.~Xiong, J.~Huang, and W.-m. Hwu, ``Deepstore: In-storage acceleration for intelligent queries,'' in \emph{Proceedings of the 52nd Annual IEEE/ACM International Symposium on Microarchitecture}, 2019, pp. 224--238.

\bibitem{gu2016biscuit}
B.~Gu, A.~S. Yoon, D.-H. Bae, I.~Jo, J.~Lee, J.~Yoon, J.-U. Kang, M.~Kwon, C.~Yoon, S.~Cho \emph{et~al.}, ``Biscuit: A framework for near-data processing of big data workloads,'' \emph{ACM SIGARCH Computer Architecture News}, vol.~44, no.~3, pp. 153--165, 2016.

\bibitem{shi2023efficient}
K.~Shi, M.~Wang, X.~Tan, Q.~Li, and T.~Lei, ``Efficient dynamic reconfigurable cnn accelerator for edge intelligence computing on fpga,'' \emph{Information}, vol.~14, no.~3, p. 194, 2023.

\bibitem{colbert2021competitive}
I.~Colbert, J.~Daly, K.~Kreutz-Delgado, and S.~Das, ``A competitive edge: Can fpgas beat gpus at dcnn inference acceleration in resource-limited edge computing applications?'' \emph{arXiv preprint arXiv:2102.00294}, 2021.

\bibitem{liu2022hybridc}
P.~Liu, Z.~Wei, C.~Yu, and S.~Chen, ``Hybridc: A resource-efficient cpu-fpga heterogeneous acceleration system for lossless data compression,'' \emph{Micromachines}, vol.~13, no.~11, p. 2029, 2022.

\end{thebibliography}

\end{document}